\documentclass[11pt,a4paper]{article}
\usepackage{graphicx}
\usepackage{booktabs} 

\usepackage{authblk}
\usepackage[hyperref]{acl2017}

\usepackage{times}
\usepackage{latexsym}
\usepackage{color}

\usepackage{multirow}

\usepackage{flushend}

\newcommand{\Ni}{({\em i})~}
\newcommand{\Nii}{({\em ii})~}
\newcommand{\Niii}{({\em iii})~}
\newcommand{\Niv}{({\em iv})~}
\newcommand{\Nv}{({\em v})~}

\newcommand{\good}{\textsc{Good}\,}
\newcommand{\bad}{\textsc{Bad}\,}
\newcommand{\potential}{\textsc{Potentially Useful}\,}

\usepackage{url}

\aclfinalcopy 

\title{Fully Automated Fact Checking Using External Sources}

\date{}
\author[1]{Georgi Karadzhov}
\author[2]{Preslav Nakov}
\author[2]{Llu\'{i}s M\`{a}rquez}
\author[2]{Alberto Barr\'on-Cede\~no}
\author[1]{Ivan Koychev}

\affil[1]{Sofia University ``St. Kliment Ohridski'', Bulgaria}
\affil[2]{Qatar Computing Research Institute, HBKU, Qatar}
\affil[ ]{\textit{georgi.m.karadjov@gmail.com, \{pnakov, lmarquez, albarron\}@hbku.edu.qa}}
\affil[ ]{\textit{koychev@fmi.uni-sofia.bg}}

\begin{document}
\maketitle
\begin{abstract}
Given the constantly growing proliferation of false claims online in recent years, there has been also a growing research interest in automatically distinguishing false rumors from factually true claims. Here, we propose a general-purpose framework for fully-automatic fact checking using external sources, tapping the potential of the entire Web as a knowledge source to confirm or reject a claim. Our framework uses a deep neural network with LSTM text encoding to combine semantic kernels with task-specific embeddings that encode a claim together with pieces of potentially-relevant text fragments from the Web, taking the source reliability into account. The evaluation results show 
good performance on two different tasks and datasets: \Ni rumor detection and \Nii fact checking of the answers to a question in community question answering forums.

\end{abstract}

\section{Introduction}
\label{sec:intro}

Recent years have seen the proliferation of deceptive information online.
With the increasing necessity to validate the information from the Internet, \emph{automatic fact checking} has emerged as an important research topic. It is at the core of multiple applications, e.g.,  discovery of fake news, rumor detection in social media, information verification in question answering systems, detection of information manipulation agents, and assistive technologies for investigative journalism. At the same time, it touches many aspects, such as credibility of users and sources, information veracity, information verification, and linguistic aspects of deceptive language. 

\noindent In this paper, we present an approach to fact-checking with the following design principles: \Ni generality, \Nii robustness, \Niii simplicity, \Niv reusability, and \Nv strong machine learning modeling. Indeed, the system makes very few assumptions about the task, and looks for supportive information directly on the Web. 
Our system works fully automatically. It does not use any heavy feature engineering and can be easily used in combination with task-specific approaches as well, as a core subsystem. Finally, it combines the representational strength of recurrent neural networks with kernel-based classification.

\begin{figure*}
\centering
\includegraphics[width=1.0\textwidth]{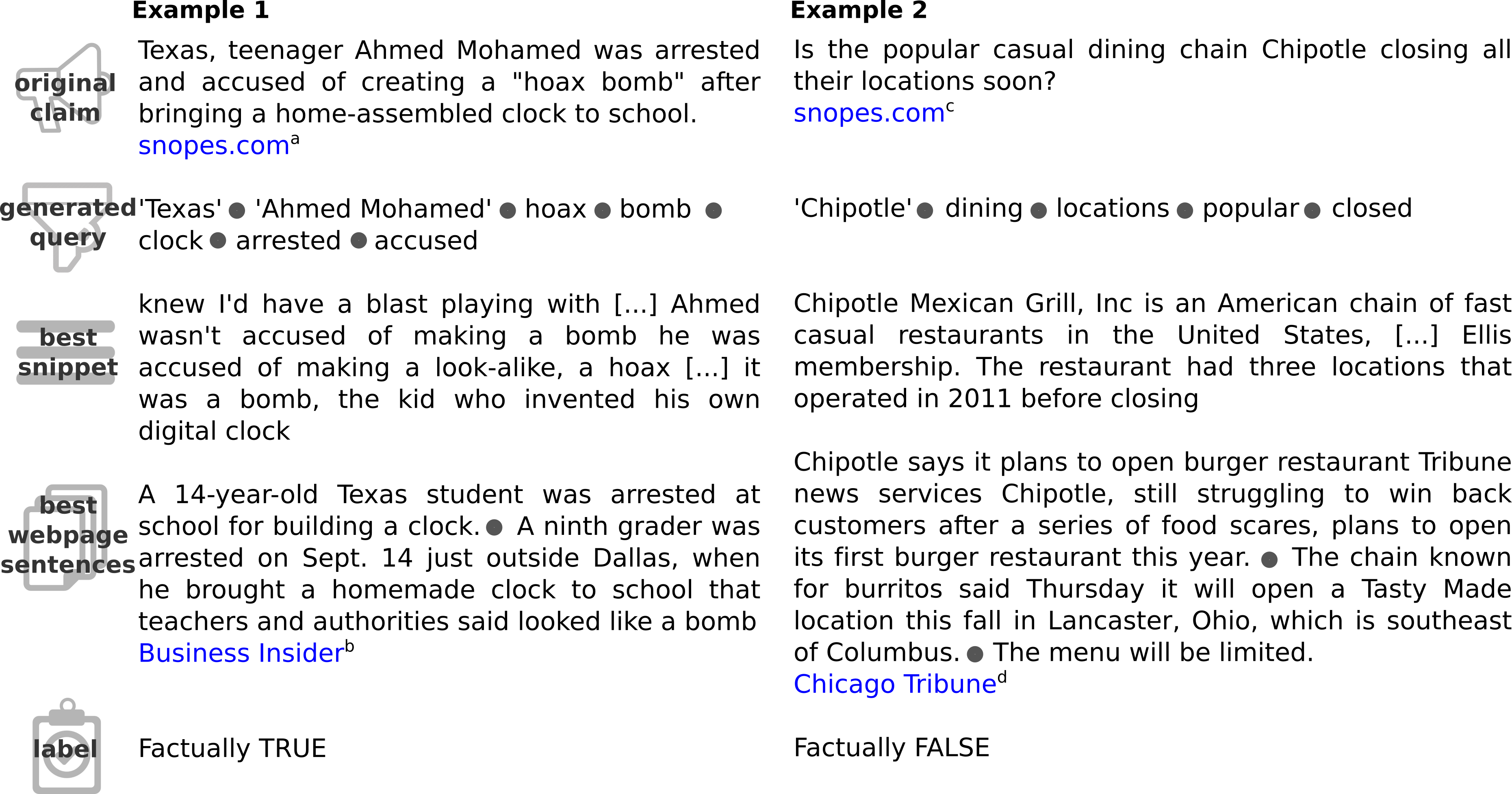}
\vspace{2mm}

\scriptsize
\begin{tabular}{p{15.8cm}} 
 $^a$\url{http://www.snopes.com/2015/09/16/ahmed-mohamed/} \\
 $^b$\url{http://www.businessinsider.com/ahmed-mohamed-arrested-irving-texas-clock-bomb-2015-9} \\
 $^c$\url{ http://www.snopes.com/chipotle-closing/} \\
 $^d$\url{http://www.chicagotribune.com/business/ct-chipotle-burger-restaurant-20160728-story.html} \\
\end{tabular}

\caption{Example claims 
and the information we use to predict whether they are factually true or false.\label{fig:example_claim}}

\end{figure*}

The system starts with a claim to verify. First, we automatically convert the claim into a query, which we execute against a search engine in order to obtain a list of potentially relevant documents. Then, we take both the snippets and the most relevant sentences in the full text of these Web documents, and we compare them to the claim. The features we use are dense representations of the claim, of the snippets and of related sentences from the Web pages, which we automatically train for the task using 
Long Short-Term Memory networks (LSTMs).
We also use the final hidden layer of the neural network as a task-specific embedding of the claim, together with the Web evidence. We feed all these representations as features, together with pairwise similarities, into a Support Vector Machine (SVM) classifier using an RBF kernel to classify the claim as True or False.

Figure~\ref{fig:example_claim} presents a real example from one of the datasets we experiment with. The left-hand side of the figure contains a True example, while the right-hand side shows a False one. We show the original claims from \url{snopes.com}, the query generated by our system, and the information retrieved from the Web (most relevant snippet and text selection from the web page). 
The veracity of the claim can be inferred from the textual information.

\noindent Our contributions can be summarized as follows:

\begin{itemize}
\item We propose a general-purpose light-weight framework for fully-automatic fact checking using evidence derived from the Web.
\item We propose a deep neural network with LSTM encoding to combine semantic kernels with task-specific embeddings that encode a claim together with pieces of potentially-relevant text fragments from the Web, taking the source reliability into account. 
\item We further study factuality in community Question Answering (cQA), and we create a new high-quality dataset, which we release to the research community. To the best of our knowledge, we are the first to study factuality of answers in cQA forums, and our dataset is the first dataset specifically targeting factuality in a cQA setting.
\item We achieve strong results on two different tasks and datasets ---rumor detection and fact checking of the answers to a question in community question answering forums---, thus demonstrating the generality of the approach and its potential applicability to different fact-checking problem formulations.
\end{itemize}


\noindent The remainder of this paper is organized as follows.
Section~\ref{sec:method} introduces our method for fact checking claims using external sources.
Section~\ref{sec:experiments} presents our experiments and discusses the results.
Section~\ref{sec:application2cqa} describes an application of our approach to a different dataset and a slightly different task: fact checking in community question answering forums.
Section~\ref{sec:related} presents related work.
Finally, Section~\ref{sec:conclusions} concludes and suggests some possible directions for future work.

\section{The Fact-Checking System}
\label{sec:method}

Given a claim, our system searches for support information on the Web in order to verify whether the claim is likely to be true. The three steps in this process are 
\Ni external support retrieval, \Nii text representation, and \Niii veracity prediction.

\subsection{External Support Retrieval}

\begin{figure*}
\centering
\includegraphics[width=0.65\textwidth]{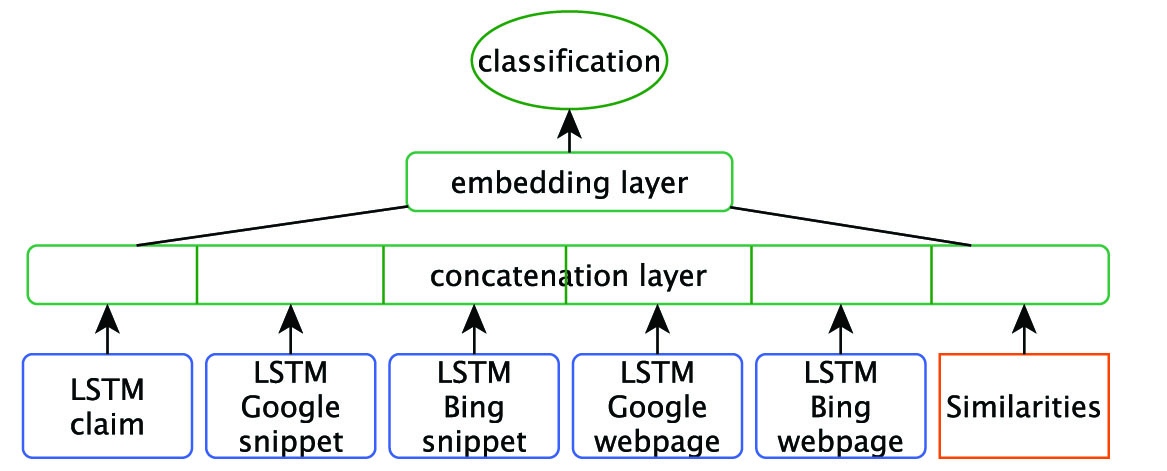}\vspace*{4mm}

\includegraphics[width=0.93\textwidth]{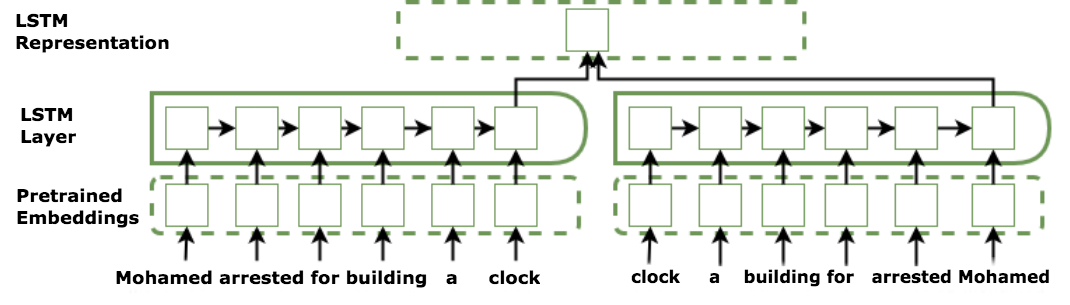}
\caption{Our general neural network architecture (top) and detailed LSTM representation (bottom).
Each blue box in the top consists of the bi-LSTM structure in the bottom.\label{fig:NN}}
\end{figure*}

This step consists of generating a query out of the claim and querying a search engine (here, we experiment with Google and Bing) in order to retrieve supporting documents. 
Rather than querying the search engine with the full claim (as on average, a claim is two sentences long), we generate a shorter query following the lessons highlighted in \cite{potthast2013overview}. 

\noindent As we aim to develop a general-purpose fact checking system, we use an approach for query generation that does not incorporate any features that are specific to claim verification (e.g., no temporal indicators).

We rank the words by means of \emph{tf-idf}. We compute the \emph{idf} values on a 2015 Wikipedia dump and the English Gigaword.\footnote{\url{catalog.ldc.upenn.edu/ldc2011t07}} \citet{potthast2013overview} suggested that a good way to perform high-quality search is to only consider the verbs, the nouns and the adjectives in the claim; thus, we exclude all words in the claim that belong to other parts of speech. Moreover, claims often contain named entities (e.g., names of persons, locations, and organizations); hence, we augment the initial query with all the named entities from the claim's text. We use IBM's AlchemyAPI\footnote{\url{www.ibm.com/watson/alchemy-api.html}} to identify named entities. Ultimately, we generate queries of 5--10 tokens, which we execute against a search engine.
We then collect the snippets and the URLs in the results, skipping any result that points to a domain that is considered unreliable.\footnote{We created such a list by manually checking the 100 most frequent domains in the results, which we accummulated across many queries and experiments.}
Finally, if our query has returned no results, we iteratively relax it by dropping the final tokens one at a time.

\subsection{Text Representation}
\label{sub:representation}

Next, we build the representation of a claim and the corresponding snippets and Web pages. 
First, we calculate three similarities (a)~between the claim and a snippet, or (b)~between the claim and a Web page: \Ni cosine with \emph{tf-idf}, \Nii cosine over embeddings, and \Niii containment~\cite{lyon2001detecting}. 
We calculate the embedding of a text as the average of the embeddings of its words; for this, we use pre-trained embeddings from GloVe \cite{pennington-socher-manning:2014:EMNLP2014}.
Moreover, as a Web page can be long, we first split it into a set of rolling sentence triplets, then we calculate the similarities between the claim and each triplet, and we take the highest scoring triplet.
Finally, as we have up to ten hits from the search engine, we take the maximum and also the average of the 
three similarities over the snippets 
and over the Web pages.

We further use as features the embeddings of the claim, of the best-scoring snippet, and of the best-scoring sentence triplet from a Web page. We calculate these embeddings \Ni as the average of the embeddings of the words in the text, and also \Nii using LSTM encodings, which we train for the task as part of a deep neural network (NN).
We also use a task-specific embedding of the claim together with all the above evidence about it, which comes from the last hidden layer of the NN.



\subsection{Veracity Prediction}


Next, we build classifiers: neural network (NN), support vector machines (SVM), and a combination thereof (SVM+NN).

\paragraph{NN.}
The architecture of our NN is shown on top of Figure~\ref{fig:NN}. We have five LSTM 
sub-networks, one for each of the text sources from two search engines: \emph{Claim}, \emph{Google Web page}, \emph{Google snippet}, \emph{Bing Web page}, and \emph{Bing snippet}. The claim is fed into the neural network as-is. As we can have multiple snippets, we only use the best-matching one as described above. Similarly, we only use a single best-matching triple of consecutive sentences from a Web page. 
We further feed the network with the similarity features described above.

\noindent All these vectors are concatenated and fully connected to a much more compact hidden layer that captures the task-specific embeddings. This layer is connected to a softmax output unit to classify the claim as true or false. 
The bottom of Figure~\ref{fig:NN} represents the generic architecture of each of the LSTM components.
The input text is transformed into a sequence of word embeddings, which is then passed to the bidirectional LSTM layer to obtain a representation for the full sequence.



 
\begin{table*}[t]
\centering
\footnotesize
\begin{tabular}{l|ccc|ccc|ccr}
\multicolumn{1}{c}{ } & \multicolumn{3}{c}{\bf False Claims} & \multicolumn{3}{c}{\bf True Claims} & \multicolumn{3}{c}{\bf Overall}\\
\bf Model & \bf P & \bf R & \bf F1 & \bf P & \bf R & \bf F1 & \bf AvgR & \bf AvgF$_1$ & \bf Acc\\
\hline
SVM + NN & 84.1 & 86.3 & 85.2 & 71.1 & 67.5 & 69.2 & \bf 76.9 & \bf 77.2 & \bf 80.0\\ 
NN & 79.6 & 92.5 &  85.5 & 77.8 & 52.5 & 62.7 & \bf 72.5 & \bf 74.1 & \bf 79.2 \\ 
SVM & 75.0 & 86.3 & 80.2 & 60.7 & 42.5 & 50.0 & \bf 64.4 & \bf 65.1 & \bf 71.7\\ 
\hline
all \emph{false} & 66.7 & 100.0 & 80.0 & -- & 0.0 & 0.0 & 50.0 &  40.0 & 66.7\\
all \emph{true}  & -- & 0.0 & 0.0 & 33.3 & 100.0 & 50.0 & 50.0 & 25.0 &  33.3\\
\hline
\end{tabular}
\caption{\label{table:resultsRumor} Results on the \emph{rumor detection dataset} using Web pages returned by the search engines.}
\end{table*} 
 
\paragraph{SVM.} 
Our second classifier is an SVM with an RBF kernel.
The input is the same as for the NN: word embeddings and similarities.
However, the word embeddings this time are calculated by averaging rather than using a bi-LSTM.

\paragraph{SVM + NN.} 
Finally, we combine the SVM with the NN by augmenting the input to the SVM with the values of the units in the hidden layer. This represents a task-specific embedding of the input example, and in our experiments it turned out to be quite helpful. Unlike in the SVM only model, this time we use the bi-LSTM embeddings as an input to the SVM. Ultimately, this yields a combination of deep learning and task-specific embeddings with RBF kernels.


\section{Experiments and Evaluation}
\label{sec:experiments}



\subsection{Dataset}
\label{sub:dataset}

We used part of the rumor detection dataset created by~\citet{ma2016detecting}. While they analyzed a claim based on a set of potentially related tweets, we focus on the claim itself and on the use of supporting information from the Web. 

\noindent The dataset consists of 992 sets of tweets, 778 of which are generated starting from a claim on \url{snopes.com}, which~\newcite{ma2016detecting} converted into a query. Another 214 sets of tweets are tweet clusters created by other researchers~\cite{Castillo:2011:ICT:1963405.1963500,Kwon:2013} with no claim behind them.
\newcite{ma2016detecting} ignored the claim and did not release it as part of their dataset. 
We managed to find the original claim for 761 out of the 778 \url{snopes.com}-based clusters.

Our final dataset consists of 761 claims from \url{snopes.com}, which span various domains including politics, local news, and fun facts. Each of the claims is labeled as factually \emph{true} (34\%) or as a \emph{false} rumor (66\%). 
We further split the data into 509 for training, 132 for development, and 120 for testing.
As the original split for the dataset was not publicly available, and as we only used a subset of their data, we had to make a new training and testing split.
Note that we ignored the tweets, as we wanted to focus on a complementary source of information: the Web.
Moreover, \newcite{ma2016detecting} used manual queries, while we use a fully automatic method.
Finally, we augmented the dataset with Web-retrieved snippets, Web pages, and sentence triplets from Web pages.\footnote{All the data, including the splits, is available at \url{github.com/gkaradzhov/FactcheckingRANLP}}

\subsection{Experimental Setup}

We tuned the architecture (i.e., the number of layers and their size) and the hyper-parameters of the neural network on the development dataset.
The best configuration uses a bidirectional LSTM with 25 units. It further uses a RMSprop optimizer with 0.001 initial learning rate, L2 regularization with $\lambda$=0.1, and 0.5 dropout after the LSTM layers. The size of the hidden layer is 60 with \emph{tanh} activations.
We use a batch of 32 and we train for 400 epochs.

For the SVM model, we merged the development and the training dataset, and we then ran a 5-fold cross-validation with grid-search, looking for the best kernel and its parameters. We ended up selecting an RBF kernel with $c=16$ and $\gamma=$0.01.

\subsection{Evaluation Metrics}

The evaluation metrics we use are P~(precision), R~(recall), and F$_1$, which we calculate with respect to the false and to the true claims. We further report AvgR (macro-average recall), AvgF$_1$ (macro-average F$_1$), and Acc (accuracy).



\subsection{Results}

Table~\ref{table:resultsRumor} shows the results on the test dataset.
We can see that both the NN and the SVM models improve over the majority class baseline (all false rumors) by a sizable margin. Moreover, the NN consistently outperforms the SVM by a margin on all measures. Yet, adding the task-specific embeddings from the NN as features of the SVM yields overall improvements over both the SVM and the NN in terms of avgR, avgF$_1$, and accuracy.
We can see that both the SVM and the NN overpredict the majority class (false claims); however, the combined SVM+NN model is quite balanced between the two classes.

Table~\ref{table:results:rumor:detailed} compares the performance of the SVM with and without task-specific embeddings from the NN, when training on Web pages vs. snippets, returned by Google vs. Bing vs. both. The NN embeddings consistently help the SVM in all cases. Moreover, while the baseline SVM using snippets is slightly better than when using Web pages, there is almost no difference between snippets vs. Web pages when NN embeddings are added to the basic SVM. Finally, gathering external support from either Google or Bing makes practically no difference, 
and using the results from both together does not yield much further improvement.
Thus, \Ni the search engines already do a good job at generating relevant snippets, and one does not need to go and download the full Web pages, and \Nii the choice of a given search engine is not an important factor. These are  good news for the practicality of our approach.

Unfortunately, direct comparison with respect to~\cite{ma2016detecting} is not possible. First, we only use a subset of their examples: 761 out of 993 (see Section~\ref{sub:dataset}),
and we also have a different class distribution. 
More importantly, they have a very different formulation of the task: for them, the claim is not available as input (in fact, there has never been a claim for 21\% of their examples); rather an example consists of a set of tweets retrieved using \emph{manually} written queries. 

\noindent In contrast, our system is fully automatic and does not use tweets at all. 
Furthermore, their most important information source is the change in tweets volume over time, which we cannot use. Still, our results are competitive to theirs when they do not use temporal features. 

\begin{table}
\centering
\footnotesize
\resizebox{\linewidth}{!}{%
\begin{tabular}{llcccc}
\bf Model & \bf External support & \bf AvgR & \bf AvgF$_1$ & \bf Acc\\
\hline
\hline
SVM + NN& Bing+Google; pages & 76.9 & 77.2 & \bf 80.0\\ 
SVM		& Bing+Google; pages & 64.4 & 65.1 & \bf 71.7\\ 
\hline
SVM + NN & Bing+Google; snippets & 75.6 & 75.6 & \bf 78.3\\ 
SVM & Bing+Google; snippets & 68.1 & 69.0 & \bf 74.2\\ 
\hline
\hline
SVM + NN & Bing; pages & 77.5 & 77.0 & \bf79.2\\ 
SVM & Bing; pages & 66.9 & 67.5 & \bf 72.5\\ 
\hline
SVM + NN & Bing; snippets & 76.3 & 76.4 & \bf 79.2\\ 
SVM & Bing; snippets & 68.8 & 69.7 & \bf 75.0\\ 
\hline
\hline
SVM + NN & Google; pages & 73.1 & 74.2 & \bf 78.3\\ 
SVM & Google; pages & 63.1 & 63.8 & \bf 71.7\\ 
\hline
SVM + NN & Google; snippets & 73.1 & 74.2 & \bf 78.3\\ 
SVM & Google; snippets & 65.6 & 66.6 & \bf 73.3\\ 
\hline
\hline
\multicolumn{2}{l}{baseline (all false claims)} & 50.0 & 40.0 & \bf 66.7\\
\hline
\end{tabular}
}
\caption{\label{table:results:rumor:detailed} Results using an SVM with and without task-specific embeddings from the NN on the \emph{Rumor detection dataset}. Training on Web pages vs. snippets vs. both.}
\end{table}

To put the results in perspective,
we can further try to make an indirect comparison to the very recent paper by \citet{Popat:2017:TLE:3041021.3055133}. They also present a model to classify true vs. false claims extracted from \url{snopes.com}, by using information extracted from the Web. Their formulation of the task is the same as ours, but our corpora and label distributions are not the same, which makes a direct comparison impossible. Still, we can see that regarding overall classification accuracy they improve a baseline from 73.7\% to 84.02\% with their best model, i.e., a 39.2\% relative error reduction. In our case, we go from 66.7\% to 80.0\%, i.e., an almost identical 39.9\% error reduction. These results are very encouraging, especially given the fact that our model is much simpler than theirs
regarding the sources of information used (they model the stance of the text, the reliability of the sources, the language style of the articles, and the temporal footprint).

\section{Application to cQA}
\label{sec:application2cqa}

Next, we tested the generality of our approach by applying it to a different setup:
fact-checking the answers in community question answering (cQA) forums. 
As this is a new problem, for which no dataset exists, we created one.
We augmented with factuality annotations the cQA dataset from SemEval-2016 Task 3 (CQA-QA-2016)~\cite{nakov-EtAl:2016:SemEval}. Overall, we annotated 249 question--answer, or $q$-$a$, pairs (from 71 threads): 
128~factually true and 121 factually false answers.

\begin{figure}
\centering
\includegraphics[width=0.40\textwidth]{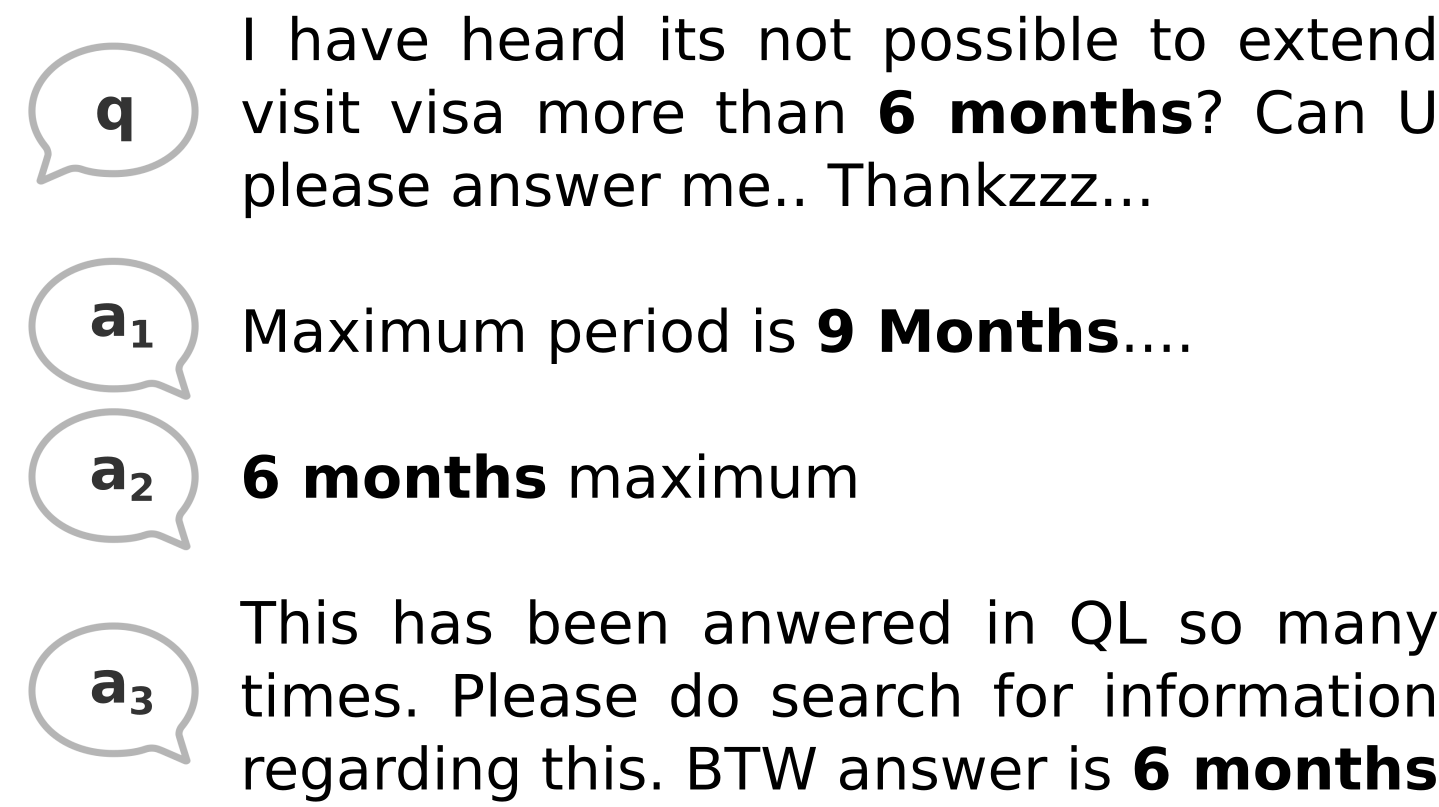}
\caption{\label{fig:example}  Example from the cQA forum dataset.}
\label{fig:example1}
\end{figure}

Each question in CQA-QA-2016 has a subject, a body, and meta information: ID, category (e.g., \emph{Education}, and \emph{Moving to Qatar}), date and time of posting, user name and ID. 
We selected 
only the factual questions such as ``\emph{What is Ooredoo customer service number?}'', thus filtering out all
(\emph{i})~socializing, e.g., ``\emph{What was your first car?}'', (\emph{ii})~requests for opinion/advice/guidance, e.g., ``\emph{Which is the best bank around??}'', and (\emph{iii})~questions containing multiple sub-questions, e.g., ``\emph{Is there a land route from Doha to Abudhabi. If yes; how is the road and how long is the journey?}''

Next, we annotated for veracity the answers to the retained questions. Note that in CQA-QA-2016, each answer has a subject, a body, meta information (answer ID, user name and ID), and a judgment about how well it addresses the question of its thread: \good\  vs.\ \potential\ vs.\ \bad. 
We only annotated the \good\ answers.\footnote{See \cite{nakov-EtAl:2017:SemEval} for an overview of recent approaches to finding \good\ answers for cQA.}
We further discarded answers whose factuality was very time-sensitive (e.g., ``\emph{It is Friday tomorrow.}'', ``\emph{It was raining last week.}'')\footnote{Arguably, many answers are somewhat time-sensitive, e.g., ``\emph{There is an IKEA in Doha.}'' is true only after IKEA opened, but not before that. In such cases, we just used the present situation as a point of reference.},
or for which the annotators were unsure. 

\begin{table}
\footnotesize
\centering
\begin{tabular}{llccc}
& \textbf{Label} & \textbf{Answers} \\
\hline
$+$ & \textsc{Factual - True} & 128 \\
\hline
$-$ & \textsc{Factual - Partially True} & 38 \\
$-$ & \textsc{Factual - Conditionally True} & 16 \\
$-$ & \textsc{Factual - False} & 22 \\
$-$ & \textsc{Factual - Responder Unsure} & 26 \\
$-$ & \textsc{NonFactual} & 19 \\
\hline
& \bf TOTAL & \bf 249 \\
& $+$ \bf \textsc{Positive} & \bf 128\\
& $-$ \bf \textsc{Negative} & \bf 121\\
\hline
\end{tabular}
\caption{Distribution of the answer labels.}
\label{table:comment-labels-distribution}
\end{table}

\begin{table*}[tbh]
\centering
\footnotesize
\begin{tabular}{l|ccc|ccc|ccr}
\multicolumn{1}{c}{ } & \multicolumn{3}{c}{\bf False Claims} & \multicolumn{3}{c}{\bf True Claims} & \multicolumn{3}{c}{\bf Overall}\\
\bf Model & \bf P & \bf R & \bf F1 & \bf P & \bf R & \bf F1 & \bf AvgR & \bf AvgF1 & \bf Acc\\
\hline
SVM + NN & 72.2 & 76.5 & 74.3 & 73.3 & 68.8 & 71.0 & \bf 72.7 & \bf 72.7 & \bf 72.7\\ 
SVM & 70.6 & 70.6 & 70.6 & 68.8 & 68.8 & 68.8 & \bf 69.7 & \bf 69.7 & \bf 69.7\\ 
NN & 61.1 & 64.7 &  62.9 & 60.0 & 56.3 & 58.1 & \bf 60.5 & \bf 60.5 & \bf 60.6 \\ 
\hline
all \emph{false} & 51.5 & 100.0 & 68.0 & -- & 0 & 0 & \bf 50.0 & \bf 34.0 & \bf 51.5\\
all \emph{true} & -- & 0.0 & 0.0 & 48.5 & 100.0 & 65.3 & \bf 50.0 & \bf 32.7 & \bf 48.5\\
\hline
\end{tabular}
\caption{\label{table:results:QL} Results on the cQA answer fact-checking problem. 
}
\end{table*}

\noindent We targeted very high quality, and thus we did not use crowdsourcing for the annotation, as pilot annotations showed that the task was very difficult and that it was not possible to guarantee that \textit{Turkers} would do all the necessary verification, e.g.,~gathering evidence from trusted sources. Instead, all examples were first annotated independently by four annotators,
%
and then \emph{each example} was discussed in detail to come up with a final label.
We ended up with 249 \textsc{Good} answers to 71 different questions, which we annotated for factuality: 128 \textsc{Positive} and 121 \textsc{Negative} examples. See  Table~\ref{table:comment-labels-distribution} for details.


We further split our dataset into 185 $q$--$a$ pairs for training, 31 for development, and 32 for testing, preserving the general positive:negative ratio, and 
making sure that the questions for the $q$--$a$ pairs did not overlap between the splits.%

Figure~\ref{fig:example1} presents an excerpt of an example from the dataset, with one question and three answers selected from a longer thread. Answer $a_1$ contains false information, while $a_2$ and $a_3$ are true, as can be checked on an official governmental website.\footnote{\url{https://www.moi.gov.qa/site/english/departments/PassportDept/news/2011/01/03/23385.html}}

We had to fit our system for this problem, as here we do not have claims, but a question and an answer. So, we constructed the query from the concatenation of $q$ and $a$.
Moreover, as Google and Bing performed similarly, we only report results using Google. We limited our run to snippets only, as we have found them rich enough above (see Section~\ref{sec:experiments}). 
Also, we had a list of reputed and Qatar-related sources for the domain, and we limited our results to these sources only. 
This time, we had more options to calculate similarities compared to the rumors dataset: we can compare against $q$, $a$, and $q$--$a$;
we chose to go with the latter.
For the LSTM representations, we use both the question and the answer.

\noindent Table~\ref{table:results:QL} shows the results on the cQA dataset. Once again, our models outperformed all baselines by a margin. This time, the predictions of all models are balanced between the two classes, which is probably due to the dataset being well balanced in general. The SVM model performs better than the NN by itself. This is due to the fact that the cQA dataset is significantly smaller than the \textit{rumor detection} one. Thus, the neural network could not be trained effectively by itself. Nevertheless, the task-specific representations were useful and combining them with the SVM model yielded consistent improvements on all the measures once again.



\section{Related Work}
\label{sec:related}

Journalists, online users, and researchers are well aware of the proliferation of false information on the Web, and topics such as information credibility and fact checking are becoming increasingly important as research directions. 
For example, there was a recent 2016 special issue of the ACM Transactions on Information Systems journal on Trust and Veracity of Information in Social Media \cite{Papadopoulos:2016:OSI}, there was a SemEval-2017 shared task on Rumor Detection \cite{derczynski-EtAl:2017:SemEval}, and there is an upcoming lab at CLEF-2018 on Automatic Identification and Verification of Claims in Political Debates \cite{RANLP2017:debates}.

The credibility of contents on the Web has been questioned by researches for a long time. 
While in the early days the main research focus was on online news portals~\cite{brill2001online,finberg2002digital,Hardalov2016}, the interest has eventually shifted towards social media \cite{Castillo:2011:ICT:1963405.1963500,PlosONE:2016,Popat:2017:TLE:3041021.3055133,RANLP2017:clickbait},
which are abundant in 
sophisticated malicious users such as opinion manipulation \emph{trolls}, paid \cite{Mihaylov2015ExposingPO} or just perceived \cite{Mihaylov2015FindingOM,mihaylov-nakov:2016:P16-2}, \emph{sockpuppets} \cite{Maity:2017:DSS:3022198.3026360}, \emph{Internet water army} \cite{Chen:2013:BIW:2492517.2492637}, and \emph{seminar users} \cite{SeminarUsers2017}.

\noindent For instance, \citet{Canini:2011} studied the credibility of Twitter accounts (as opposed to tweet posts), and found that both the topical content of information sources and social network structure affect source credibility.
Other work, closer to ours, aims at addressing credibility assessment of rumors on Twitter as a problem of finding false information about a newsworthy event~\cite{Castillo:2011:ICT:1963405.1963500}. This model considers
user reputation, writing style, and various time-based features, among others.

Other efforts have focused on news communities. For example, several truth discovery algorithms are 
combined in an ensemble method for veracity estimation in the VERA system~\cite{Ba:2016:VERA}. They proposed a platform for end-to-end truth discovery from the Web: extracting unstructured information from multiple sources, combining information about single claims, running an ensemble of algorithms, and visualizing and explaining the results. 
They also explore two different real-world application scenarios for their system:
fact checking for crisis situations and evaluation of trustworthiness of a rumor. However, the input to their model is structured data, while here we are interested in unstructured text as input.

Similarly, the task defined by~\citet{mukherjee2015leveraging} combines three objectives: assessing the credibility of a set of posted articles, estimating the trustworthiness of sources, and predicting  user's expertise.
They considered a manifold of features characterizing language, topics and Web-specific statistics (e.g., review ratings) on top of a continuous conditional random fields model.
In follow-up work, \citet{popat2016credibility} proposed a model to support or refute claims from \url{snopes.com} and Wikipedia 
by considering supporting information gathered from the Web. They used the same task formulation for claims as we do, but different datasets.
In yet another follow-up work, \newcite{Popat:2017:TLE:3041021.3055133} proposed a complex model that considers stance, source reliability, language style, and temporal information.

\noindent Our approach to fact checking is related: we verify facts on the Web. However, we use a much simpler and feature-light system, and a different machine learning model.
Yet, our model performs very similarly to this latter work (even though a direct comparison is not possible as the datasets differ), which is a remarkable achievement given the fact that we consider less knowledge sources, we have a conceptually simpler model, and we have six times less training data than \newcite{Popat:2017:TLE:3041021.3055133}.

Another important research direction is on using tweets and temporal information for checking the factuality of rumors.
For example, \citet{Ma:2015:DRU} used temporal patterns of rumor dynamics to detect false rumors and to predict their frequency.
\citet{Ma:2015:DRU} focused on detecting false rumors in Twitter using time series.
They used the change of social context features over a rumor's life cycle in order to detect rumors at an early stage after they were broadcast.

A more general approach for detecting rumors is explored by~\citet{ma2016detecting}, who used recurrent neural networks to learn hidden representations that capture the variation of contextual information of relevant posts over time. Unlike this work, we do not use microblogs, but we query the Web directly in search for evidence. Again, while direct comparison to the work of \citet{ma2016detecting} is not possible, due to differences in dataset and task formulation, we can say that our framework is competitive when temporal information is not used. More importantly, our approach is orthogonal to theirs in terms of information sources used, and thus, we believe there is potential in combining the two approaches.




In the context of question answering, there has been work on assessing the credibility of an answer, e.g., based on intrinsic information~\cite{banerjee-han:2009:NAACLHLT09-Short}, i.e., without any external resources. In this case, the reliability of an answer is measured by computing the divergence between language models of the question and of the answer.
The spawn of community-based question answering Websites also allowed for the use of other kinds of information. Click counts, link analysis (e.g., PageRank), and user votes have been used to assess the quality of a posted answer~\cite{Agichtein:2008:FHC:1341531.1341557,Jeon:2006:FPQ:1148170.1148212,Jurczyk:2007:DAQ:1321440.1321575}. Nevertheless, these studies address the answers' credibility level just marginally.

\noindent Efforts to determine the credibility of an answer in order to assess its overall quality required the inclusion of content-based information~\cite{Su-EtAl:2010:PACLIC2010}, e.g., verbs and adjectives such as \emph{suppose} and \emph{probably}, which cast doubt on the answer.
Similarly, \citet{lita2005qualitative}
used source credibility (e.g., does the document come from a government Website?), 
sentiment analysis, and answer contradiction compared to other related answers. 

Overall, \emph{credibility} assessment for question answering has been mostly modeled at the feature level, with the goal of assessing the quality of the answers. A notable exception is the work of \cite{RANLP2017:credibility:trolls}, where credibility is treated as a task of its own right.
Yet, note that \emph{credibility} is different from \emph{factuality} (our focus here) as the former is a subjective perception about whether a statement is credible, rather than verifying it as true or false as a matter of fact; still, these notions are often wrongly mixed in the literature.
To the best of our knowledge, no previous work has targeted fact-checking of answers in the context of community Question Answering by gathering external support.



\section{Conclusions and Future Work}
\label{sec:conclusions}

We have presented and evaluated a general-purpose method for fact checking that relies on retrieving supporting information from the Web and comparing it to the claim using  machine learning. Our method is lightweight in terms of features and can be very efficient because it shows good performance by only using the snippets provided by the search engines. The combination of the representational power of neural networks with the classification of kernel-based methods has proven to be crucial for making balanced predictions and obtaining good results. 
Overall, the strong performance of our model across two different fact-checking tasks confirms its generality and potential applicability for different domains and for different fact-checking task formulations.

In future work, we plan to test the generality of our approach by 
applying it to these and other datasets in combination with complementary methods, e.g., those focusing on microblogs and temporal information in Twitter to make predictions about rumors~\cite{Ma:2015:DRU,ma2016detecting}.
We also want to explore the possibility of providing justifications for our predictions, and we plan to integrate our method into a real-world application.

\section*{Acknowledgments}
This research was performed by the Arabic Language Technologies group at Qatar Computing Research Institute, HBKU\@, within the Interactive sYstems for Answer Search project ({\sc Iyas}).

\bibliographystyle{acl_natbib}
\bibliography{acl2017} 
\end{document}